\documentclass{article}
\usepackage{ijcai21}

\usepackage{times}
\usepackage{soul}
\usepackage{url}
\usepackage[hidelinks]{hyperref}
\usepackage[utf8]{inputenc}
\usepackage[small]{caption}
\usepackage{graphicx}
\usepackage{amsmath}
\usepackage{amsthm}
\usepackage{booktabs}
\usepackage{algorithm}
\usepackage{algorithmic}
\usepackage{amssymb}
\usepackage{multirow}
\usepackage{array}
\usepackage{makecell}

\newcommand{\bW}{\mathbf{W}}

\newcommand{\bc}{\mathbf{c}}

\newcommand{\bh}{\mathbf{h}}

\newcommand{\bo}{\mathbf{o}}

\newcommand{\by}{\mathbf{y}}
\newcommand{\bz}{\mathbf{z}}
\newcommand{\ie}{\textit{i.e.}}
\newcommand{\eg}{\textit{e.g.}}
\title{Emotion Eliciting Machine: Emotion Eliciting Conversation Generation\\ based on Dual Generator}

\author{
Hao Jiang$^1$\and
Yutao Zhu$^2$\and
Xinyu Zhang$^1$\and
Zhicheng Dou$^3$\and
Pan Du$^2$\and
Te Pi$^1$\And
Yantao Jia$^1$
\affiliations
$^1$Huawei Poisson Lab., Hangzhou, Zhejiang, China \\
$^2$Université de Montréal, Montréal, Québec, Canada \\
$^3$Gaoling School of Artificial Intelligence, Renmin University of China, Beijing, China
\emails
\{jianghao66,zhangxinyu35,pite,jiayantao\}@huawei.com, \\
\{yutao.zhu,pan.du\}@umontreal.ca,
dou@ruc.edu.cn
}
\begin{document}

\maketitle

\begin{abstract}
Recent years have witnessed great progress on building emotional chatbots. Tremendous methods have been proposed for chatbots to generate responses with given emotions. However, the emotion changes of the user during the conversation has not been fully explored. In this work, we study the problem of positive emotion elicitation, which aims to generate responses that can elicit positive emotion of the user, in human-machine conversation.
%This task aims at generating responses for the user input to elicit positive emotion of the user. 
We propose a weakly supervised Emotion Eliciting Machine (EEM) to address this problem.
Specifically, we first collect weak labels of user emotion status changes in a conversion based on a pre-trained emotion classifier. Then we propose a dual encoder-decoder structure to model the generation of responses in both positive and negative side based on the changes of the user's emotion status in the conversation. An emotion eliciting factor is introduced on top of the dual structure to balance the positive and negative emotional impacts on the generated response during emotion elicitation. The factor also provides a fine-grained controlling manner for emotion elicitation.
Experimental results on a large real-world dataset show that EEM outperforms the existing models in generating responses with positive emotion elicitation.

% Building emotional chatbots has great potential value for human beings, which draws increasing attention in recent years. While most existing approaches focus on generating responses with specified emotions, they might cause offending or unexpected responses in certain context. Instead, we study the problem of response generation which aims to improve emotional experience of users. The problem is more challenging since the generated responses should not only be coherent in content, but also consider users' emotional status. Towards this goal, we propose the Emotion Eliciting Machine (EEM) that models the conversation generation conditioned on the emotional dynamics of users. EEM is fully data-driven that it can be developed on large-scale conversation logs without manual labeling. In applications, EEM can fine-grained control the positivity of emotion elicitation of the generated responses. Experimental results show that EEM outperforms the existing models in emotion eliciting conversation generation with its superior controllability and flexibility.
\end{abstract}

\section{Introduction}
Building an emotional intelligent agent, which can perceive, integrate, understand, and regulate emotions, is one of the ultimate goals of artificial intelligence~\cite{salovey1990emotional,picard2000affective}. Particularly, emotional chatbots are attracting more and more research interests due to their wide application in intelligent assistant, emotional escort, and online consultation~\cite{zhou2018emotional,huang2018automatic,shantala2018neural,zhou-wang-2018-mojitalk,colombo-etal-2019-affect,shen-feng-2020-cdl}. 

% Due to the accelerating pace of life and work, people in modern society are under increasing psychological pressure. Meanwhile, more and more children and the elderly are in lack of companionship and in desperate need of psychological comfort. Therefore, emotional chatbots can be helpful if they can interact with users with emotional behaviors, such as comfort, encouragement, and appreciation. Compared with human emotional service, emotional chatbots are always on call with very low cost.

\begin{figure}[t!]
    \centering
    \includegraphics[width=\linewidth]{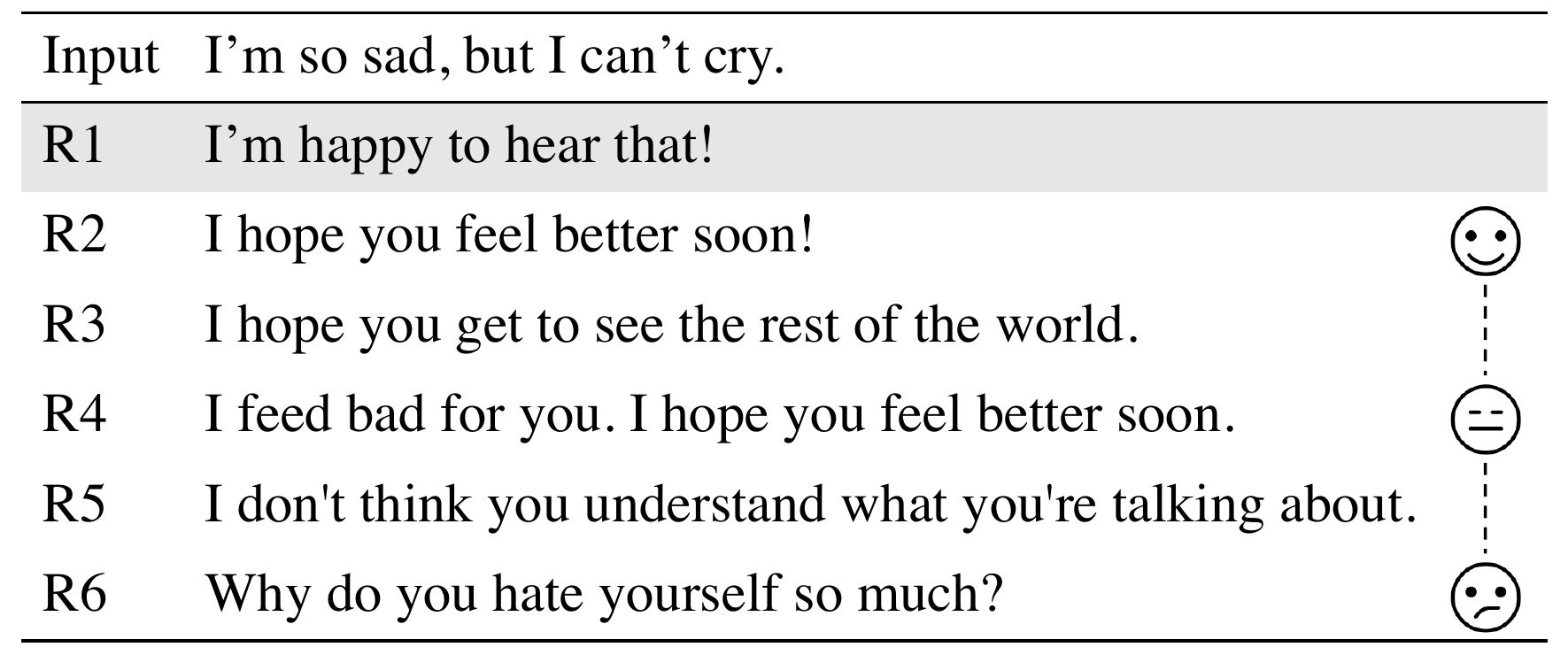}
    \caption{An example of positive emotion elicitation in conversation. R1 is a response that neglects the user's emotion. R2-R6 are several responses with different granularities (from most positive to less positive) of positive emotion elicitation.}
    \vspace{-5px}
    \label{fig:example}
\end{figure}

Existing studies on emotional chatbots can be roughly categorized into two groups. The first group of methods usually require chatbots to generate responses with the predefined emotion~\cite{zhou2018emotional,huang2018automatic,zhou-wang-2018-mojitalk,shen-feng-2020-cdl}, while the current emotion status of the user that the machine is chatting with is neglected during the response generation. 
% Though more emotional responses can be generated with the help of the given emotion labels, the emotion status of the user is not considered. 
As a result, if a user worries about a problem they met, such as the example shown in Figure~\ref{fig:example}, a chatbot with the given emotion label ``happy'' may generate a response like ``I'm happy to hear that!'', which may make the user feel uncomfortable. This example shows that a response with the positive emotion (such as R1) may fail to make the user happier. To solve this problem, the other group of approaches focus on modeling the emotion change (reaction) of the user rather than merely controlling the emotion of responses generated by the chatbots~\cite{hasegawa-etal-2013-predicting,lubis2018eliciting,lubis2019positive}. These methods aim at eliciting positive emotion of the user through the conversation\footnote{The model can also be used to elicit the negative emotion of the user, but it is less applied in practice.}. For example, in Figure~\ref{fig:example}, R2-R4 are some typical generated responses that can bring the user with positive emotions to some extent. We study the latter problem in this work -- how to generate responses to elicit positive emotions of the user.

Hasegawa et al.~\shortcite{hasegawa-etal-2013-predicting} leverage a statistical translation model for emotion eliciting in human-machine conversation. The model is initially trained on a general-purpose conversation corpus, then fine-tuned on datasets with emotion labels. Lubis et al.~\shortcite{lubis2018eliciting,lubis2019positive} trained a hierarchical recurrent neural network to generate emotional responses based on a positive-emotion eliciting corpus. With manually collected labels about emotions, these methods have enabled the emotion eliciting capability of the chatbots successfully.
However, there are two major drawbacks in the existing methods. First, most existing methods rely on positive emotion labels. Negative samples are usually neglected based on an assumption that they are useless for training an emotion eliciting model~\cite{hasegawa-etal-2013-predicting,lubis2018eliciting,lubis2019positive}. We consider that the emotions expressed by the human utterances are often too subtle to be expressed in merely with positive emotion labels, \eg, as the complicated expression ``\textit{I'm so sad, but I can't cry.}'' demonstrated in Figure~\ref{fig:example}. Correspondingly, an ideal emotion eliciting bot should be able to adapt to subtle expressions by balancing the positive and negative feelings to avoid some extreme mistakes, such as responding ``\textit{I'm happy to hear that!}'' to the above expression. Second, the emotion labels are often collected manually in the existing methods, which is quite labor-intensive and thus makes them less practical, especially when the training corpus is large.

% \cite{hasegawa-etal-2013-predicting} first focus on the task of eliciting a specific category of emotion in conversation with translation-based response generators. \cite{lubis2018eliciting,lubis2019positive} pay more attention to user's emotional experience and propose an emotion-sensitive model to generate responses that can elicit positive emotion of the user, but the model needs to be trained on positive-emotion eliciting corpus. 
% On the other hand, manually pre-processing corpus to be positive-eliciting on a large scale is very expensive. 
% It is also a waste of data resources to give up all the dialogues with neutral or negative emotion elicitation since these dialogues are still helpful for modeling semantics and common logic.

% Few studies have explored the affects of the generated responses on users' emotional status. The pioneering work \cite{hasegawa-etal-2013-predicting} tag the emotion of the addressees, and trains a machine translation model on the tagged corpus to generate responses. \cite{lubis2018eliciting,lubis2019positive} propose a sequence-to-sequence (seq2seq) response generator from positive-emotion eliciting responses collected through crowd-sourcing. However, these ?????? approaches are prone to suffer from the problem of data sparseness. Because only few utterances in a conversation corpus are emotion-tagged, and constructing such a corpus is expensive

To address these issues, in this paper, we propose a weakly supervised \textbf{Emotion Eliciting Machine} (EEM) to generate responses that adaptively elicit positive emotions. Specifically, we \textbf{first} leverage an emotion change classifier to generate weak supervision signals based on the predicted emotion scores of the utterances in the dialogue session. We leverage the difference of the emotion score between two consecutive utterances of a user to measure the effect of emotion elicitation of the utterance made by the other user. More specifically, supposing ($U_1$, $R_1$, $U_2$) is a triplet of consecutive utterances, $U_1$ and $U_2$ are utterances of a user, and $R_1$ is the response made by another user to $U_1$, the difference of the emotion scores between $U_1$ and $U_2$ is used to indicate the effect of emotion elicitation of $R_1$. 
With this method, we can generate a large number of weakly labeled samples for our model training\footnote{Our experiments demonstrate that these weak labels have high consistency with human labeling.} and save lots of human labours. 
With these generated labels, we \textbf{further} introduce an emotion eliciting factor on top of a dual-generator structure to generate responses that can balance positive and negative feelings according to the emotion changes. The positive generator aims to emphasize positive emotions during the response generation process, while the negative generator focuses on the negative ones. Our method learns to dynamically balance the influences of the positive and negative generators by an emotion eliciting factor during the training procedure. In such a way, we can fully utilize all utterances for training a chatbot, and at the same time, we can learn 
a fine-grained emotion elicitation model. 
%Secondly, we propose to compute an emotion eliciting factor based on the emotion changes (\ie, the sentiment score difference) to model the fine-grained eliciting process. We design a dual structure to generate a response in both positive and negative side.
%During training process, our method can dynamically tune the influence of the positive or negative generator to model the effect of elicitation in different granularities (with different emotion eliciting factors). 
In the inference stage, the emotion eliciting factor can be arbitrarily set to control the positivity of emotion elicitation in the generated responses. Experimental results on a large-scale real human conversation corpus (Reddit) show that our method can be well-trained with the obtained weak labels and outperforms existing state-of-the-art methods on the task of emotion elicitation.  

% To this end, we propose Emotion Eliciting Machine (EEM) that can be constructed on general conversation datasets without additional manual pre-processing or labels of emotion elicitation. To model different emotional impacts (i.e., positive and negative) of a response, we introduce a complementary encoder-decoder model, which integrates a positive generator and a negative generator to generate a response. 

The main contributions of this work are as follows:

(1) We propose EEM for emotion eliciting conversation generation, which can make use of not only positive emotion samples as in the literature, but also the negative ones, so as to void generating positive responses in inappropriate contexts.

(2) We propose two novel mechanisms to improve the adaptability and controllability of the model: 1) computing an emotion eliciting factor that balances the positive and negative emotional impacts of a response; 2) a dual structure that models the generation of responses in both positive and negative sides to obtain a dynamic controlling of the positivity of emotion elicitation.

(3) The results on the large scale Reddit conversation corpus show that EEM outperforms the existing models in both automatic and human evaluation measures.

\section{Related Work}
\noindent\textbf{Open-domain Chatbots and Response Generation}
Existing methods on building open-domain chatbots can be categorized into two groups: retrieval-based and generation-based.
Retrieval-based methods try to find the most reasonable response from a large repository of conversational data~\cite{lowe-etal-2015-ubuntu,wu-etal-2017-sequential,yuan-etal-2019-multi}.
On the other hand, generation-based methods are mainly based on the sequence-to-sequence (Seq2seq) architecture with attention mechanism and aim at generating a new response for conversation context~\cite{sordoni2015neural,li2015diversity,niu-bansal-2018-polite,baheti-etal-2018-generating}. 
In this study, we focus on the generation-based chatbots modeling. 

% Our work falls into the group of researches building chatbots with end-to-end neural networks which draw increasing interest in recent years. Most of these works are inspired by the success of seq2seq framework \cite{sutskever2014sequence} which is widely used for sequence generation conditioned on another sequence. Typically, an RNN is used as an encoder to encode the previous sentence in a conversation, and another RNN is utilized to predict the response \cite{oriol2015neural,sordoni2015neural,niu-bansal-2018-polite,baheti-etal-2018-generating}.
% Compared with previous retrieval-based or machine-translation based approaches, the end-to-end one requires much fewer hand-crafted rules or tuning steps, and can generalize better to unseen context.

\noindent\textbf{Emotional Chatbots}
% One drawback of the standard seq2seq model is that the generated response is completely dependent on the training corpus. As the result is not controllable, it is risky to generate unexpected or meaningless responses. Moreover, in most applications it is expected that the agent generates responses with some requirements. We refer them as controllable conversation models.
% \cite{li-etal-2016-persona} introduce a speaker vector into the input of decoder to generate personal-based responses.
% \cite{xing2017topic} propose a topic aware seq2seq model to generate responses consistent with the topics of previous message. The key part of the model is an attention mechanism that summarizes the context of input message and its topic words. \cite{choudhary2017domain} propose the domain seq2seq model to generate domain-specific responses, by using a domain classifier for re-ranking the responses.
Emotional chatting model is a special type of controllable conversation models, which aims to control the emotion express of responses. Existing methods can be divided into two categories according to their different purposes: (1) Some methods focus on controlling the emotion of chatbots. For example, Zhou et al.~\shortcite{zhou2018emotional} proposed an emotional chatting model with neural networks, which takes the target emotion category of responses as an additional input of the decoder, and then generates emotional responses with the proposed internal and external emotion memory. 
% Huang et al.~\shortcite{huang2018automatic} proposed two methods to embedded emotion in seq2seq, either by concatenating an emotion token to the input message, or injecting the emotion into decoder. Song et al.~\shortcite{song2019generating} designed an emotional dialogue system that can express the given emotion either explicitly or implicitly, using an emotion classifier and a lexicon-based attention mechanism.
% \cite{colombo-etal-2019-affect} use a continuous representation of emotions to guide the sampling of emotion words in a seq2seq model, and apply an affective re-ranking module to improve the emotional content in generated responses. 
Shen and Feng~\shortcite{shen-feng-2020-cdl} designed a dual task to generate emotional responses and emotional queries alternatively and improved the content consistency in emotion-controllable response generation. These methods can control the chatbot to generate responses with pre-defined emotion, but they neglect the influence of the generated response on user's emotion. (2) Other methods pay less attention to the emotion of the chatbot but consider the emotional influence of the generated responses on the user, which is usually called \textit{emotion eliciting} models.
% Emotion eliciting chatbot goes further by taking the emotional influence of the generated responses into account.
Hasegawa et al.~\shortcite{hasegawa-etal-2013-predicting} first studied the task of eliciting a specific emotion in conversation. They trained a machine translation model on utterances exhibiting emotion elicitation for each emotion category. Lubis et al.~\shortcite{lubis2018eliciting,lubis2019positive} proposed an emotion-sensitive dialogue model which associates the emotion label of the target sentence for decoding. This model is more effective than the traditional hierarchical recurrent encoder-decoder~\cite{serban2016building} when trained on a manually-collected corpus containing responses of positive emotion elicitation. However, as the data with human annotation is hard to collect, their method has less application in practice.

\begin{figure*}[!tbp]
	\centering
	\includegraphics[width=0.75\linewidth]{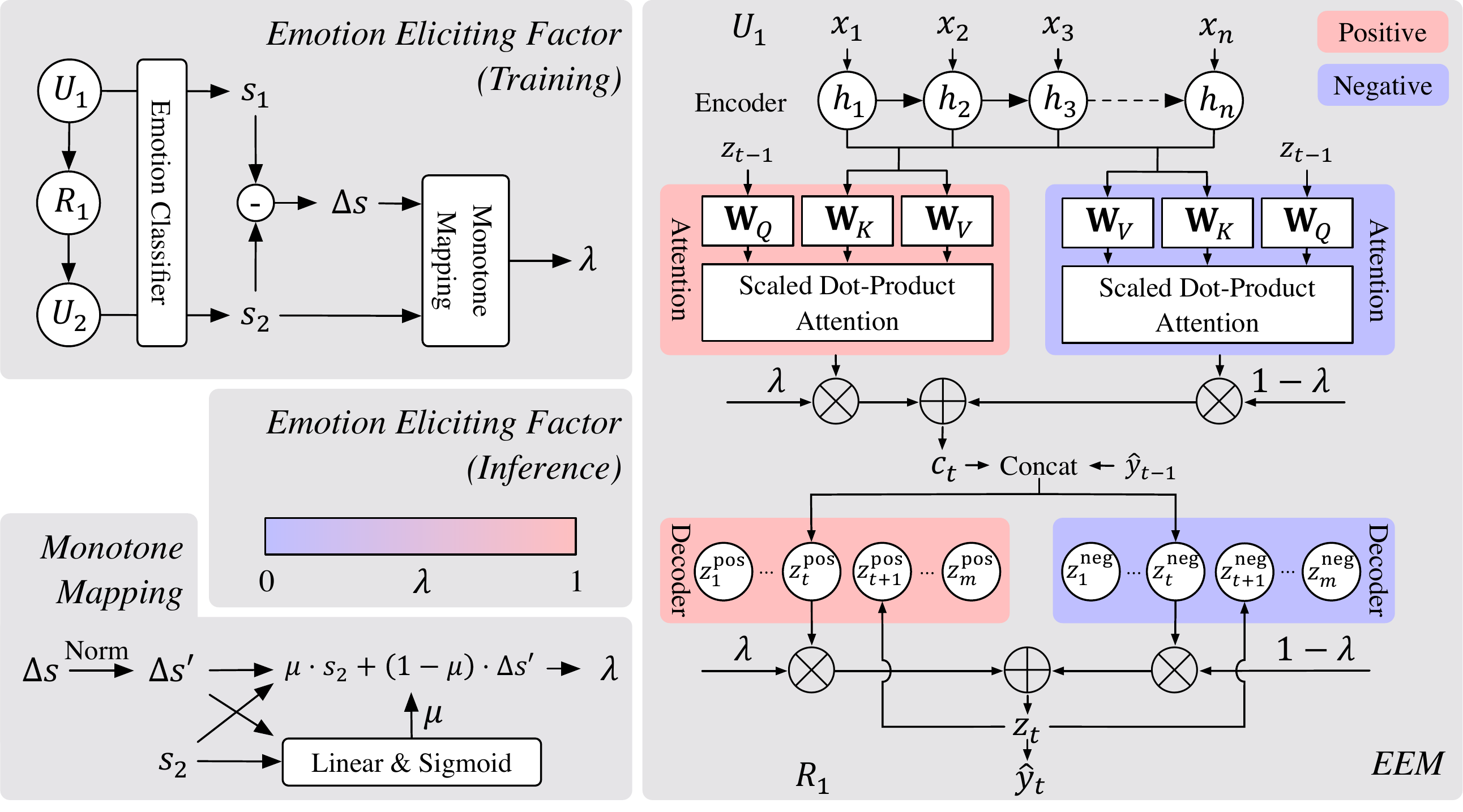}
	\caption{Overview of the proposed Emotion Eliciting Machine (EEM). 
% 	It uses two attention modules and two decoder modules to model the positive and negative emotion elicitation separately. An emotion eliciting coefficient ($\lambda$) is then used to integrate both positive and negative modules. In the stage of training, $\lambda$ is calculated from emotional dynamics of dialogue triplets to ensure that each module models pertinently (i.e., the modules activated by $\lambda$ model the pattern of positive emotion elicitation, and those activated by $1-\lambda$ model the pattern of negative emotion elicitation, respectively). In inference, $\lambda$ can be manually set up to fine-grained control the emotion eliciting effects of the generated responses.
	}
	\vspace{-5px}
	\label{pipeline}
\end{figure*}
Our method is different from existing emotion eliciting models. At first, we propose a distant labeling method to get the weak label which indicates the emotion changes of the user. Our model can be trained with such data, thus is free with human annotation. On the other hand, we compute an eliciting factor based on the value of the weak label. The eliciting factor can reflect the degree of emotion changes, which boosts our model in fine-grained emotion elicitation.
% only sensitive to emotion but not controllable. That is, the polarity and the degree of emotional elicitation of responses generated by this model depends on the distribution of training corpus. When the model is trained on general or even negative-eliciting corpus, the generated responses may be not positive-eliciting anymore.

\section{Proposed Method}
\subsection{Problem Formulation and the Overall Model}\label{sec:length}
Emotion eliciting chatbots aim at generating a response to improve the emotion status of the user (positive emotion elicitation). Specifically, we consider a dialogue with $N$ pairs of utterances $(U_1, R_1,\cdots, U_i, R_i, \cdots,U_N,R_N)$, where $U_i$ is the input message given by the user at the $i$-th turn, and $R_i$ is the corresponding response generated by the chatbot. 
% The process of controllable response generation can be considered as the probability estimation of a generated response by a language model, \ie, $P(R_{n}|U_{n}, C)$, where $U_{n}$ is the user utterance (i.e., a sequence of words) at turn $n$, and $R_{n}$ is the agent response at turn $n$. $C$ is the controllable condition.
% In this work, we expect a model to generate responses that can improve users' emotional status, which is a special task of controllable response generation.
We note $S=\{s_{i}|i\in[1,N]\}$ as the emotional status sequence of the user in the dialogue, where $s_{i}\in[0,1]$ denotes the positivity of the user's emotion at the $i$-th turn. 
% We utilize an sentiment classifier to determine $s_{n}$ based on the utterance $U_{n}$, i.e., $s_{n}=f(U_{n})$. 
%In our case, we just use the sentiment classifier off-the-shelf.

Then, the task of positive-emotion eliciting response generation can be formulated as: generating the response $R_i$ to improve the user's emotional positivity $s_{i+1}$ based on the input message $U_i$ and previous user status $s_i$. 
% \begin{align}
% 		R_{i} & \sim P(R_{i}|U_{i}, s_{i}),\label{eq:target} \\
% 		s_{i+1} & \sim (R_i, U_{i+1}),\label{eq:process} \\
% 		& \max s_{i+1}.\label{eq:max_emo}
% \end{align}
% Intuitively, our model is designed to 
% Specifically, this paper aims to build a model that can generate responses improving users' emotional positivity, given his/her previous utterance. 
To simplify the notation, we use triplet $\{U_1, R_1, U_2\}$ to denote a consecutive utterance in the dialogue, where $U_1$ is the input message from a user, $R_1$ is the corresponding response from the chatbot, and $U_2$ is the next input message/reaction from the user. 
% Comparing the emotional positivities of $U_1$ and $U_2$ can reflect the emotion changes after the user get the response $R_1$. 
%Our model is trained on these triplets, which is described in the following sections.
% We collect all the triplets as our training corpus.
% Dialogue triplets with the format of $\{U_1, R_1, U_2\}$ are utilized as the training units.

%\section{Proposed Method}
%In order to utilize the abundant open-source general dialogue data to train a positive-emotion eliciting model, we propose the Complementary Encoder-Decoder (ComEnDec) framework with a complementary coefficient $\lambda$. Figure~\ref{pipeline} shows our pipeline.

In this paper, we propose leveraging all triplets (not just triplets with $s_2 > s_1$) for training the response generation model. We introduce an emotion eliciting factor on top of a dual-generator structure to generate responses that can balance positive and negative feelings according to the emotion changes. The positive generator aims to emphasize positive emotions during the response generation process, while the negative generator focuses on the negative ones. Our method learns to dynamically balance the influences of the positive and negative generators by an emotion eliciting factor during the training procedure

The detailed structure of the proposed method is shown in Figure~\ref{pipeline}. At first, we compute an emotion eliciting factor $\lambda$ to capture the emotion changes after the user receives the response $R_1$ (Section~\ref{sec_lambdaDef}). In the training stage, this factor is computed based on the emotion score of $U_1$ and $U_2$. In the test stage, as $U_2$ is unavailable, $\lambda$ can be set arbitrarily to achieve positive emotion elicitation in different degrees. After capturing the emotion changes, we design a dual structure to model the response generation process in both positive and negative side. This structure contains an encoder with dual attentions and a dual decoder. The influence of both attention and decoder is determined by $\lambda$ so as to provide fine-grained emotion elicitation (Section~\ref{sec_gene}).
In the remaining part of the section, we will introduce each component in detail.

\subsection{Emotion Eliciting Factor $\mathbf{\lambda}$}\label{sec_lambdaDef}
%To model the fine-grained effect of the emotion elicitation, we propose an emotion eliciting factor $\lambda\in\left[0,1\right]$. 
The emotion eliciting factor $\lambda\in\left[0,1\right]$ is a continuous variable that represents the degree of positive emotion elicitation, where 1 indicates the most positive, and 0 is the most negative. Intuitively, the emotion eliciting factor is relevant to: (1) the user's emotion status before receiving $R_1$, namely $s_1$ (the status of $U_1$); and (2) the changes of the user's emotion status after receiving $R_1$, namely ($s_2-s_1$). 

The computation of the factor $\lambda$ relies on the changes of the user's emotion status, which is represented by the emotion scores' difference of the user's uterrances caused by the chatbot's response. 
Specifically, we apply a pre-trained emotion polarity classifier to automatically obtain the emotional positiveness scores $s_1,s_2\in\left[0,1\right]$ of $U_1$ and $U_2$ to represent the emotion status, respectively. Then $\lambda$ can be computed as:
\begin{align}\label{lambda}
	\lambda&=\mu\cdot s_2+(1-\mu)\cdot {\Delta s}', \\
	{\Delta s}'&=\frac{s_2 - s_1 + 1}{2}\in\left[0,1\right],
\end{align}
where the weight $\mu$ is learned by a feed-forward neural network on $s_2$ and ${\Delta s}'$ to model the implicit interactions between the emotional characteristics of $U_1$ and $U_2$ as:
\begin{align}\label{mu}
	\mu=\sigma(w_{1}s_2+w_{2}{\Delta s}' + b),
\end{align}
where $\sigma\left(\cdot\right)$ is the sigmoid function. $w_1$, $w_2$, and $b$ are parameters. 
The computation process of $\lambda$ is shown in the left side of Figure~\ref{pipeline}. In the training stage, it is determined by the emotional status $s_1$ and $s_2$, which can be obtained by the pre-trained emotion classifier. By this means, we save lots of resources on human labeling. Our experiments (Section~\ref{exp:sentlabel}) will show that these weak labels are highly consistent with human labels, which guarantees the performance of our model. In the inference stage, $\lambda$ can be arbitrarily given to control the polarity and degree of emotion elicitation. For example, to achieve the most effective positive-emotion elicitation in practical applications, we can set $\lambda=1$.

\subsection{Dual Generator}
\label{sec_gene}
The value of $\lambda$ reflects the degree of positive emotion elicitation in $R_1$. We design a dual structure to learn the generation of $R_1$ from both positive and negative side with the help of $\lambda$. For example, a more positive emotion eliciting sample (with larger $\lambda$) will contribute more on optimizing the positive generator while the less positive ones have more influence on the negative generator. In such a dual structure, all samples with different $\lambda$s can be used for training, which makes full use of the whole dataset. On the other hand, by tuning the generators in both sides, our model can achieve fine-grained emotion elicitation that is adaptive to the user's input message.

\paragraph{Encoder with Dual Attention}
During chatting, people usually pay attention to different context when constructing their messages. Therefore, we propose a two-headed attention mechanism to capture the different attention patterns from both positive and negative elicitation.

% \begin{figure}[t]
% 	\centering
% 	\includegraphics[width=0.99\linewidth,trim={5.5cm 5.5cm 5.5cm 6.5cm},clip]{attention.pdf} % Reduce the figure size so that it is slightly narrower than the column. Don't use precise values for figure width.This setup will avoid overfull boxes.
% 	\caption{Structure of the dual attention.}
% 	\label{fig_comAtt}
% \end{figure}

We adopt the attention function from Transformer \cite{vaswani2017attention} for the computational efficiency:
% where the scaling factor can alleviate the gradient vanishing problem:
\begin{align}
	e^{\left(\text{mode}\right)}_{t,j}&=\frac{1}{\sqrt{d_h}} {\left(\bW^{\left(\text{mode}\right)}_{K}\bh_j\right)}^\top \bW^{\left(\text{mode}\right)}_{Q} \bz_{t-1}, \\
	\alpha_{t,j}^{\left(\text{mode}\right)} &= \text{softmax} \left(\left[\cdots,e^{\left(\text{mode}\right)}_{t,k}, \cdots\right] \right)_{k=j}, \\
	\bc_t^{\left(\text{mode}\right)}&=\sum_{j=1}^{n}{\alpha_{t,j}^{\left(\text{mode}\right)}\bW^{\left(\text{mode}\right)}_{V}\bh_j},
\end{align}
where $\bh$ and $\bz$ are the outputs of the encoder and the decoder respectively; $\bW^{\left(\text{mode}\right)}_{\{K,V\}}\in\mathbb{R}^{d_h\times d_h}$ and $\bW^{\left(\text{mode}\right)}_{Q}\in\mathbb{R}^{d_h\times d_z}$ are parameters; $d_h$ and $d_z$ are the dimensions of $\bh$ and $\bz$; $\text{mode}\in\left\{\text{pos}, \text{neg}\right\}$; $\bc_t^{\left(\text{pos}\right)}$ and $\bc_t^{\left(\text{neg}\right)}$ are the acquired context vector by positive attention and negative attention in the time step of $t$; and $n$ is the number of tokens in the input message.

Then the two context vectors are combined based on the emotion eliciting factor ($\lambda$) to get the integrated vector:
\begin{align}\label{ct}
	\bc_t=\lambda \bc_t^{\left(\text{pos}\right)}+\left(1-\lambda\right)\bc_t^{\left(\text{neg}\right)}.
\end{align}

\paragraph{Dual Decoder}
In addition to different attentions on the context, the emotional effect of an utterance is dependent on the interaction of both the emotionally-positive pattern and the emotionally-negative pattern in the user's mind. Activations of the positive and the negative thoughts of people are diverse due to the variety of their characteristics.
% We notice that the variety of people's responses are influenced by not only different attention but also different ways of thinking. During a conversation, the emotional effect of a person's response is dependent on the interaction of the emotionally-positive pattern and the emotionally-negative pattern in his/her mind. 
% Activations of the positive and the negative thoughts of people are diverse due to the variety of people's characteristics and the dialogue context.

% \begin{figure}[t]
% 	\centering
% 	\includegraphics[width=0.99\linewidth,trim={8cm 6.5cm 8.5cm 6.5cm},clip]{decoder.pdf} % Reduce the figure size so that it is slightly narrower than the column. Don't use precise values for figure width.This setup will avoid overfull boxes.
% 	\caption{Structure of the dual decoder.}
% 	\label{fig_compDec}
% \end{figure}

To mimic such mechanism, we also apply a dual structure in the decoder side. Our dual decoder consists of two parallel RNN-based decoders, one of which (referred to as positive decoder) can model the emotionally improving pattern, and the other (referred to as negative decoder) is to model the emotionally suppressing pattern when generating a response. Concretely, the two decoders share the same network structure and the same inputs:
\begin{align}
	\label{ztk}\bz_t^{\text{pos}}&=\text{GRU}_{\text{de}}^{\text{pos}}\left(\bz_{t-1},[{\hat {\by}_{t-1}}; \bc_{t}]\right), \\
	\label{ztr}\bz_t^{\text{neg}}&=\text{GRU}_{\text{de}}^{\text{neg}}\left(\bz_{t-1},[{\hat {\by}_{t-1}}; \bc_{t}]\right),
\end{align}
where [;] is the concatenation operation, $\bc_{t}$ is obtained from Eq. (\ref{ct}), and $\hat {\by}_{t-1}$ is the embedding vector of the target word at the time step $t-1$. Finally, we use the combination of outputs from both decoders to predict the word according to the eliciting factor $\lambda$:
\begin{align}\label{zt}
	\bz_t&=\lambda \bz_t^{\text{pos}}+(1-\lambda)\bz_t^{\text{neg}}, \\
	y_t &\sim \bo_t=\text{softmax}(\bW_o \bz_t),\label{zt_2}
\end{align}
where the $\bW_o$ is the output projection matrix.

% Note that for decoding the word at each time step $t$, the outputs of the positive decoder and the negative decoder are combined based on the condition variable $\lambda$:
% $\lambda\in\left[0, 1\right]$:
%\noindent where a bigger $\lambda$ (closer to $1$) means a more positive emotion elicitation. The condition variable $\lambda$ is determined by the emotional status of $U_1$ and $U_2$ in a training triplet, which is described in Section \ref{sec_lambdaDef}.

% Then the probability distribution of the output $o_t$ is decoded from $z_t$ as:
% \begin{equation}\label{zt_2}
% 	y_t \sim o_t=\text{softmax}(W_o z_t)
% \end{equation}
% \noindent where the $W_o$ is the output projection matrix.

As seen in Eq. (\ref{ztk}) - (\ref{ztr}), the integrated hidden state of the dual decoder $\bz_{t-1}$ is fed into both the positive and the negative decoders for the next time-step computation. Since the two decoders share the same inputs at each time step, the difference of their outputs only originates from the different values of their trainable parameters. By this means, the two decoders are forced to capture the discriminant emotional patterns imposed by the controlling condition $\lambda$, respectively.

\section{Experiments}
\subsection{Dataset and Preprocessing}
\label{exp:sentlabel}
We conduct experiments on the large scale Reddit dataset\footnote{\url{https://github.com/PolyAI-LDN/conversational-datasets/tree/master/reddit}}, which contains 3.7 billion conversation comments with dialogue context \cite{henderson-etal-2019-repository}. We extract and preprocess 55.5 million $\left(U_1, R_1, U_2\right)$ dialogue triplets from the dataset after discarding the triplets which have sentences containing more than 20 words or containing non-ASCII chars. Then we filter out web symbols, normalize the punctuations, and format all the letters as lowercase.

\paragraph{Emotion Labeling}
% In this research, a sentiment classifier is needed to pseudo-label the sentiment scores of our triplet dataset, following \cite{zhou2018emotional,song2019generating}. 
Following previous approaches~\cite{zhou2018emotional,song2019generating}, we train a classifier to get the emotion scores of all triplet in the dataset. We leverage the pre-trained DeepMoji model as the classifier and fine-tune it on the SS-Twitter dataset, which is manually labeled with emotion categories~\cite{felbo-etal-2017-using}.

% The distribution of sentiment scores is shown in Figure \ref{ratio}. As can be seen, the sentiment distribution of the dataset tends to be negative. 
% After labeled, the triplet dataset is randomly split into training set, validation set, and test set by the ratio of 0.8:0.1:0.1.

% Additional evaluation is carried out to estimate the accuracy of the sentiment classifier on our conversation dataset. We randomly sampled 200 utterances for evaluation, and the sentiment distribution of the test set is consistent with the whole dataset by a KL-divergence = 0.023. Each utterance is labeled as ``positive", ``neutral" or ``negative" according to its sentiment polarity by 5 annotators. The Fleiss' kappa coefficient \cite{fleiss1971measuring} of the annotations is 0.595, indicating ``Moderate agreement".

% Since the sentiment score predicted by the classifier is continuous, we discrete it into ``positive" (score between 0 and 0.4), ``neutral" (score between 0.4 and 0.6) and ``negative" (score between 0.6 and 1) for matching with the annotation. As a result, the sentiment classifier achieves a mean accuracy = 0.688 in the evaluation.

To evaluate the performance of the classifier, we conduct a manual evaluation to estimate the accuracy of the emotion classifier on our conversation dataset, and the classifier achieves an accuracy of 0.71 (see more details in Appendix). Then, the obtained classifier is used to score each utterance in all triplets. All the triplets are randomly split into training, validation set, and test set with the ratio of 8:1:1.

% 200 utterances are randomly sampled for evaluation, and the emotion distribution is consistent with the whole dataset (the KL-divergence is 0.023). Each utterance is labeled as ``positive'', ``neutral'' or ``negative'' according to its emotion polarity by 5 annotators. Since the emotion score predicted by the classifier is continuous, we discrete it into ``negative'' (score between 0 and 0.35), ``neutral'' (score between 0.35 and 0.65), and ``positive'' (score between 0.65 and 1). As a result, the emotion classifier achieves 0.71 accuracy on average. The Fleiss' kappa coefficient~\cite{fleiss1971measuring} of the 5 annotations is 0.592, indicating ``Moderate agreement''. 

% The obtained classifier is used to score each utterance in all triplets. Then all triplets are randomly split into training, validation set, and test set with the ratio of 8:1:1. 

% The distribution of sentiment scores is shown in Figure~\ref{ratio}.

% {\color{red}However, thanks to our framework, we can still use this dataset to train a positive-emotion eliciting model. See Section~\ref{sec:exp} for more details.}

% \begin{figure}[t]
% 	\centering
% 	\includegraphics[width=0.99\linewidth,trim={2.5cm 10cm 3cm 10cm},clip]{hist.pdf}
% 	\caption{Sentiment score distribution of the Reddit dataset.}
% 	\label{ratio}
% \end{figure}

\subsection{Evaluation Metrics}
We adopt both automatic and human evaluation metrics to verify the effectiveness of our proposed method.

\noindent\textbf{(1) Automatic Evaluation}

\noindent\textbf{Quality of response}: on the content level, we adopt \textbf{perplexity}(PPL) for evaluating the quality of generated responses. This metric is recommended and widely used to evaluate generative dialogue systems~\cite{Pietquin2013survey,zhou2018emotional,lubis2018eliciting}. It is worth noting that reference-based evaluation metrics such as BLEU or embedding similarity are not applicable for our experiments because we seek to generate the response with the most positive elicitation which may be different from the original ones.

%Automatic evaluations are applied to all models\footnote{}. 
% \noindent\textbf{Perplexity:} In this work, we adopt perplexity for the evaluation of models on the content level. This metric is recommended and widely used to evaluate generative dialogue systems \cite{Pietquin2013survey,zhou2018emotional,lubis2018eliciting}.

%\noindent\textbf{Perplexity} (PPL):
% \noindent\textbf{Emotion Eliciting Impact:} For the level of emotion elicitation, an indirect method is adopted for the automatic evaluation. Firstly, we build a standard Hierarchical Recurrent Encoder-Decoder (HRED) model \cite{sordoni2015hierarchical,serban2016building} to obtain a multi-turn chatbot. The model is trained on the test set of dialogue triplets to prevent EEM from obtaining leaky information in the process of evaluations.

\begin{figure}[t]
	\centering
	\includegraphics[width=0.99\linewidth]{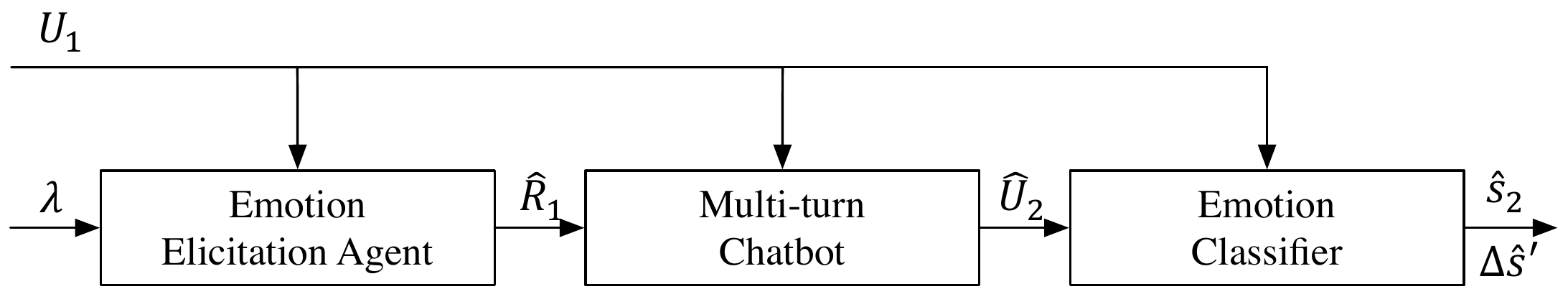} % Reduce the figure size so that it is slightly narrower than the column. Don't use precise values for figure width.This setup will avoid overfull boxes.
	\caption{Automatic evaluation of emotion elicitation}
	\vspace{-5px}
	\label{Auto_eval}
\end{figure}

\noindent\textbf{Emotion Eliciting Impact}: 
As there is no existing automatic metric for evaluating emotion elicitation, we propose an indirect approach for the evaluation, which is shown in Figure~\ref{Auto_eval}. The basic idea is to measure the user's emotion changes after receiving the response given by the chatbot. 
To achieve this, we first train a standard Hierarchical Recurrent Encoder-Decoder (STD-HRED) model \cite{sordoni2015hierarchical,serban2016building} by using the message-response reversely (namely using the original response as the input and the original message as target)\footnote{The model is trained on the test set of dialogue triplets to avoid leaky information in the process of evaluations.}. After collecting the responses $\hat R_{1}$ generated by all models, we use it as input to generate the next utterance $\hat U_{2}$ by STD-HRED with $U_{1}$ and $\hat R_{1}$ as inputs. Then we regard $\hat U_{2}$ as a probable reaction of the user after receiving the response $\hat R_{1}$, and use the pre-trained emotion classifier to compute the emotion score $\hat s_{2}$ of $\hat U_{2}$. Finally, the normalized increment $\Delta \hat{s}'$ between the emotion scores of $\hat U_{2}$ and $U_{1}$ can be calculated as well. 
%We can predict the positivty of $\hat U_{t+1}$ as $\hat s_{t+1}=f(\hat{U_{t+1}})$ using the pretrained sentiment classifier. 
Since $\hat s_{2}$ and $\Delta \hat{s}'$ can be the estimations of $s_{2}$ and $\Delta s'$, they can be used as the performance indicators of emotion elicitation.

\noindent\textbf{(2) Human Evaluation}

Due to the complexity and diversity of human language, following existing works~\cite{lubis2018eliciting,lubis2019positive}, we also conduct a human evaluation. Specifically, we randomly sample 200 utterances from the test set. For each utterance, we collect the generated responses from all models. Then we hire five experienced annotators to label them. The annotators are asked to score the generated responses by stating their agreement on two statements: 1) \textbf{Content score} ($s_c$) states the response is appropriate and natural to the post and is likely to be produced by a human. 2) \textbf{Emotional eliciting score} ($s_e$) states the response elicits a positive emotion. Rating scales of both criteria are 0, 1, and 2, which indicate disagreement, partly agreement and agreement, respectively\footnote{The Fleiss' kappa coefficients for the content score and emotional impact score are 0.473 and 0.631, indicating ``Moderate agreement" and ``Substantial agreement", respectively.}.

\subsection{Baselines}\label{sec:exp}
We compare our method with the following baselines:
% \begin{itemize}
%     \item EncDec~\cite{sutskever2014sequence}                 : A traditional Encoder-Decoder model with attention mechanism. There is no extra emotional elicitation module injected. 
%     \item Emo-HRED~\cite{lubis2018eliciting,lubis2019positive}: The state-of-art model for eliciting positive emotion in conversation. 
%     \item Emb\_$s_2$ and Emb\_$\Delta s'$: Naive baselines combining the EnDec model with the embedded representations of $s_2$ and $\Delta s'$, respectively. The embeddings serve as an extra input of the decoder to influence the generation process.
% \end{itemize}

\noindent (1) EncDec~\cite{sutskever2014sequence}: A traditional Encoder-Decoder model with attention mechanism. There is no extra emotional elicitation module injected. 

\noindent (2) Emo-HRED~\cite{lubis2018eliciting,lubis2019positive}: The state-of-art model for eliciting positive emotion in conversation. It is based on the structure of the HRED, with a latent emotion variable to be emotion-sensitive.

\noindent (3) Emb\_$s_2$ and (4) Emb\_$\Delta s'$: Naive baselines combining the EnDec model with the embedded representations of $s_2$ and $\Delta s'$, respectively. The embeddings serve as an extra input of the decoder to influence the generation process.

% \begin{figure}[t]
% \centering
% \includegraphics[width=0.9\columnwidth]{baseline} % Reduce the figure size so that it is slightly narrower than the column. Don't use precise values for figure width.This setup will avoid overfull boxes.
% \caption{The structure of the baseline model BEPM}
% \label{baseline}
% \end{figure}

\begin{table}[t]
	\centering
% 	\def\arraystretch{1.0}
	%	\begin{tabular}{C{2.5cm}|C{1.6cm}C{1.6cm}|C{1.6cm}C{1.6cm}C{1.6cm}}
	%    \begin{tabular}{p{2.5cm}<{\centering}|p{1.6cm}<{\centering}p{1.6cm}<{\centering}|p{1.6cm}<{\centering}p{1.6cm}<{\centering}p{1.6cm}<{\centering}}
	%	\begin{tabular}{c@{\hspace{1.12cm}}|c@{\hspace{0.8cm}}c@{\hspace{0.8cm}}|c@{\hspace{0.8cm}}c@{\hspace{0.8cm}}c@{\hspace{0.8cm}}}
	\setlength\tabcolsep{6pt}{
	\begin{tabular}{lccccc}
	\toprule
		& \multicolumn{3}{c}{Automatic} & \multicolumn{2}{c}{Human} \\ 
		\cmidrule(lr){2-4}\cmidrule(lr){5-6}
		Model & PPL & $\hat{s}_2$ & $\Delta \hat{s}'$ & $s_{c}$ & $s_{e}$\\
		\midrule
		EnDec & 48.94 & 0.383 & -0.027 & 1.839 &1.057 \\
		\quad on positive & 51.19 & 0.452 & 0.042 & 1.612 & 1.324 \\
		Emo-HRED & 47.00 & 0.387 & -0.023 & 1.738 & 1.132 \\
		\quad on positive & 47.66 & 0.458 & 0.047 & 1.673 & 1.348 \\
		Emb\_$s_2$ & 48.79 & 0.481 & 0.071 & 1.797 &1.359 \\
		Emb\_$\Delta s'$ & 48.87 & 0.460 & 0.050 & 1.823 & 1.365 \\
		\midrule
		EEM ($\lambda=1$) & 46.78 & \textbf{0.574} & \textbf{0.164} & \textbf{1.840} &\textbf{1.615}  \\
		\midrule
		$\lambda=s_2$ & 48.04 & 0.500 & 0.090 & 1.745 & 1.447 \\
		$\lambda=\Delta s'$ & \textbf{46.72} & 0.543 & 0.133 & 1.765 &1.515  \\
		\textit{w/o} Dual Attn. & 46.77 & 0.541 & 0.131 & 1.633 & 1.474  \\
		\textit{w/o} Dual Dec. & 48.81 & 0.479 & 0.069 & 1.763 & 1.415  \\
		\bottomrule
	\end{tabular}
	}
	% \caption{Manual evaluation with scores of grammar ($s_g$), relevance ($s_r$) and emotion elicitation ($s_e$). Automatic evaluation with perplexity ($ppl$), $\hat{s}_2 $ and $\Delta \hat{s}'$.}
	\caption{Evaluation results of all the models.}
	\vspace{-5px}
	\label{table2}
\end{table}

\begin{figure}[t]
	\centering
	\includegraphics[width=0.9\linewidth,trim={0.4cm 0.77cm 0.5cm 0.5cm},clip]{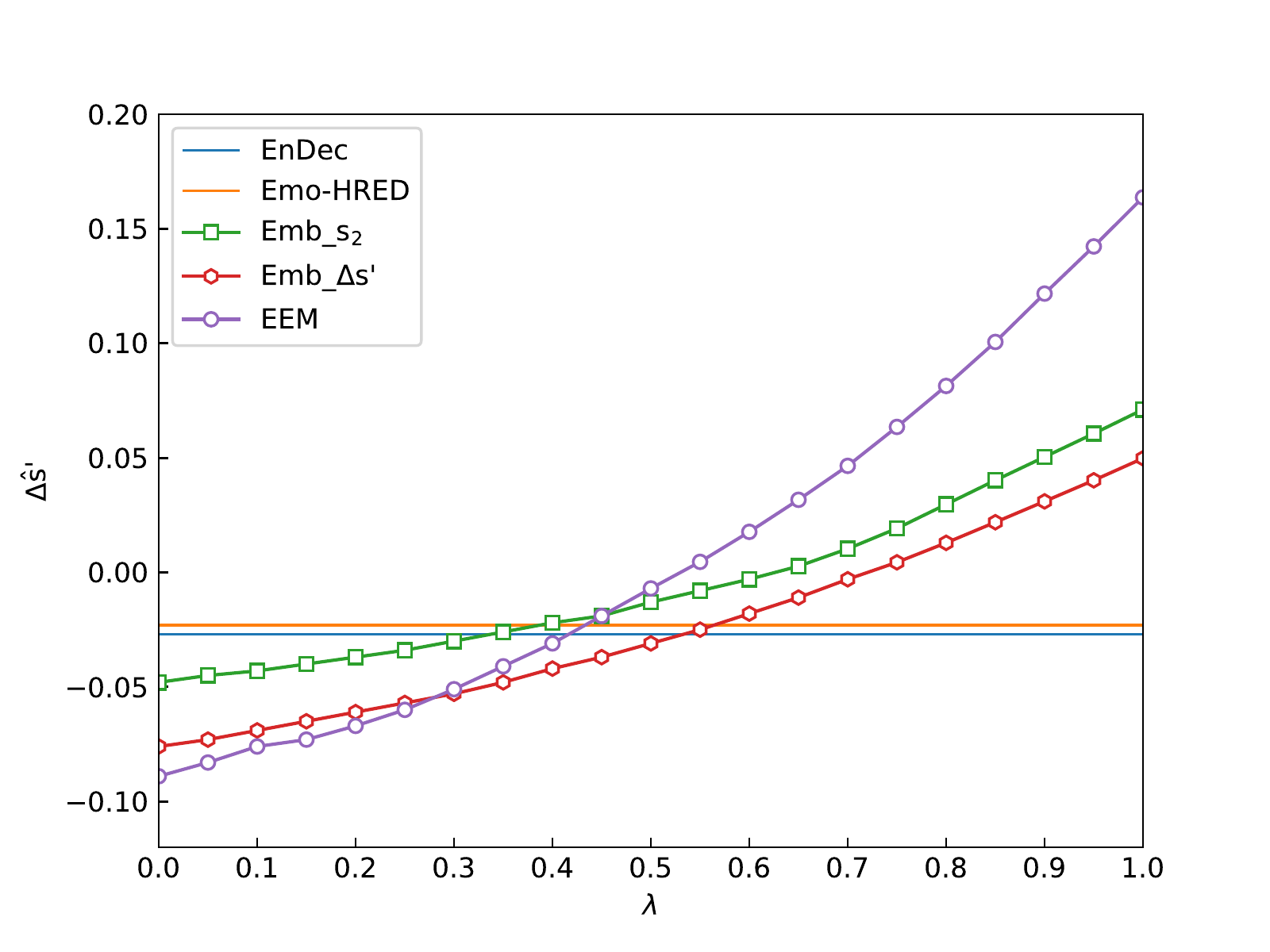}
%\vspace*{2mm}
	\caption{Curves of $\Delta \hat{s}'$ with different emotion eliciting factors.}
	\vspace{-5px}
	\label{lamda_curve}
\end{figure}

% 0.75
% \newcommand{\tabincell}[2]{\begin{tabular}{@{}#1@{}}#2\end{tabular}}
% \begin{table*}[t]
% 	\centering
% 	\small
% 	%\setlength{\tabcolsep}{3.1mm}{
% 	\begin{tabular}{p{3.5cm}|p{1.7cm}<{\centering}|p{1.7cm}<{\centering}|p{8.2cm}} %l(left)居左显示 r(right)居右显示 c居中显示
% 		\toprule
% 		\multicolumn{1}{p{3.0cm}|}{User Input}&\multicolumn{2}{c|}{Model}&\multicolumn{1}{c}{Response}\\
% 		\midrule
% 		\multirow{8}{3.0cm}{I'm so sad, but I can't cry.} & \multicolumn{2}{c|}{EnDec} & I don't think you know what the word means.\\ 
% 		\cmidrule{2-4}
% 		& \multicolumn{2}{c|}{Emo-HRED} & Me too, but I don't want to cry.\\ 
% 		\cmidrule{2-4}
% 		& \multirow{6}{*}{\tabincell{c}{EEM}}&$\lambda=1.0$ & \textbf{I hope you feel better soon!}\\
% 		& & $\lambda=0.8$ & I hope you get to see the rest of the world.\\ 
% 		& & $\lambda=0.6$ & I feel bad for you. I hope you feel better soon.\\ 
% 		& & $\lambda=0.4$ & I don't think you know what you're talking about.\\ 
% 		& & $\lambda=0.2$ & I don't think you understand what you're talking about.\\ 
% 		& & $\lambda=0.0$ & Why do you hate yourself so much?\\
% 		\bottomrule
% 	\end{tabular}
% 	%}
% 	\caption{Sample responses generated by EnDec, Emo-HRED, and EEM with different conditions.}
% 	\label{table5}
% \end{table*}

\newcommand{\tabincell}[2]{\begin{tabular}{@{}#1@{}}#2\end{tabular}}
\begin{table}[t]
	\centering
	\small
	\begin{tabular}{p{0.55cm}<{\centering}|p{0.3cm}<{\centering}|p{6.2cm}} %l(left)居左显示 r(right)居右显示 c居中显示
		\toprule
		\multicolumn{2}{c|}{User Input}
		& I'm so sad, but I can't cry. \\
		\midrule
% 		\multicolumn{2}{c|}{Model}&\multicolumn{1}{c}{Response}\\
% 		\midrule
		\multicolumn{2}{c|}{EnDec} & I don't think you know what the word means.\\ 
		\cmidrule{1-3}
		\multicolumn{2}{c|}{Emo-HRED} & Me too, but I don't want to cry.\\ 
		\cmidrule{1-3}
		\multirow{6}{*}{\tabincell{c}{EEM\\($\lambda$)}}& 1.0 & \textbf{I hope you feel better soon!}\\
		& 0.8 & I hope you get to see the rest of the world.\\ 
		& 0.6 & I feel bad for you. I hope you feel better soon.\\ 
		& 0.4 & I don't think you know what you're talking about.\\ 
		& 0.2 & I don't think you understand what you're talking about.\\ 
		& 0.0 & Why do you hate yourself so much?\\
		\bottomrule
	\end{tabular}
	%}
	\caption{Sample responses generated by EnDec, Emo-HRED, and EEM with different emotion eliciting factors.}
	\vspace{-5px}
	\label{table5}
\end{table}

\subsection{Overall Results}
% For evaluate models' performance of positive-emotion elicitation, condition variable $\lambda$ of Emb\_$s_2$, Emb\_$\Delta s'$ and ablation models is set to 1.0 when generating responses.  
% Quantitative experiments are carried out to evaluate different models' performance of generation and positive-emotion elicitation. 
To maximize the effect of positive elicitation in the generated responses, the value of $\lambda$ in Emb\_$s_2$, Emb\_$\Delta s'$, and all ablation models is set as 1.0 in the test stage.

% The Fleiss' kappa coefficient is adopted to assess the consistency of the inter-rater agreement. For the content score and emotional impact score, the corresponding kappa coefficients are 0.464 and 0.647, indicating ``Moderate agreement" and ``Substantial agreement", respectively.

The experimental results are shown in the upper side of Table~\ref{table2}. We can see: (1) EEM outperforms all the baselines significantly in terms of positive-emotion elicitation in both automatic and human evaluations. Specifically, EEM outperforms the previous best baseline by 0.114 on both $\hat s_{2}$ and $\Delta \hat{s}'$. This indicates that EEM can improve the user's emotion status by generating responses with the most positive effect of elicitation. (2) The values of $\Delta \hat{s}'$ obtained by both EnDec and Emo-HRED are negative. The reason is that both of these two methods are originally designed for datasets with only positive eliciting samples. Such datasets heavily rely on human annotation, which is often unavailable in real scenario.
(3) To mimic a similar training process, we also train EnDec and Emo-HRED on the positive subset of the data (\ie, $\Delta s'>0$ and $s_2>0.5$). The corresponding experimental results are shown in Table~\ref{table2} with an additional mark ``on positive". We can observe that the models can also achieve positive emotion elicitation ($\Delta \hat{s}'>0$). However, there is still a large gap between these models and our EEM. This proves that: a) Our proposed emotion eliciting factor can reflect the different degrees of emotion elicitation, thus help the model to achieve fine-grained elicitation and make full use of the dataset; b) Training only on positive eliciting dataset is insufficient for building a positive emotion eliciting chatbot and a well-designed structure is necessary for this task. (4) The effect of positive emotion elicitation by Emb\_$s_2$ and Emb\_$\Delta s'$ are very limited. This demonstrates that simply integrating the emotion information into the model is insufficient for generating desirable responses. Similar problems are also reported by~\cite{zhou2018emotional}. (5) According to the human evaluation results, EEM also has superiority on the positive emotion elicitation. Interestingly, the general quality of the generated responses (measured by $s_e$) is also improved. This inspires us that considering the user's emotion changes is an effective way to improve the chatbots. 

% Meanwhile, EEM outperforms EnDec and Emo-HRED in terms of both the emotion scores, $s_e$ and $\Delta \hat{s}'$. 
% The results show: 1) Injecting emotion information is helpful for emotion elicitation in response generation. 2) 

% On the other hand, $s_e$ and $\Delta \hat{s}'$ of Emo-HRED show a slight improvement compared with EnDec's scores, but underperform all the other models conditioned on external emotion information. It confirms 

% Moreover, we conduct ablation studies to investigate the influence of the dynamic calculation of $\lambda$, the Complementary Attention and the Complementary Decoder separately with the following models:

% \begin{itemize}
% 	\item $\lambda\_s_2$/$\lambda\_\Delta s'$: EEM with $\lambda$ simplified to $s_2$ and $\Delta s'$ in the training procedure, respectively.
% 	\item w/o CA: EEM with Complementary Attention replaced by a traditional attention.
% 	\item w/o CD: EEM with Complementary Decoder replaced by a traditional decoder.
% \end{itemize}

\subsection{Ablation Study}
We investigate the effect of each module in our model in this section. The results are shown in the lower side of Table~\ref{table2}.

First, we simplify $\lambda$ in EEM to $s_2$ or $\Delta s'$, respectively. The performance degradation demonstrates that the emotion eliciting effect is related to both the status of $U_2$ and the emotion changes $\Delta s$, which is consistent with our assumption. In addition, compared with the model using $s_2$ or $\Delta s'$ as additional input (\ie, Emb\_$s_2$ and Emb\_$\Delta s'$), EEM achieves better results. This confirms the superiority of our proposed dual structures in emotion elicitation. 

Second, we replace the dual attention and the dual decoder by a traditional attention and a tradition decoder respectively. These two models are denoted as ``\textit{w/o} Dual Attn.'' and ``\textit{w/o} Dual Dec.''. Experimental results show that both of the dual structures are helpful in improving performance. Specifically, the dual decoder brings more than 14\% improvements in terms of emotion elicitation under both automatic and human evaluation. \textbf{This demonstrates the effectiveness of our proposed dual structure}.

Moreover, we conduct additional experiments to investigate the emotion eliciting performance of these models when given different emotion eliciting factors (\ie, $s_2$ for Emb\_$s_2$, $\Delta s'$ for Emb\_$\Delta s'$ and $\lambda$ for EEM) for the generation. As shown in Figure~\ref{lamda_curve}, the curves of Emb\_$s_2$, Emb\_$\Delta s'$ and EEM show a positive correlation between $\Delta \hat{s}'$ and the factor. In general, when given a bigger factor, these models tend to generate more positive eliciting responses accordingly. Among all the models, EEM shows the highest responsivity between $\Delta \hat{s}'$ and the factor, which proves that \textbf{EEM is more controllable and flexible in fine-grained emotion elicitation}.

% Besides, the ablation models including w/o CD and w/o CA also perform well. Emb shows a slight improvement compared with EnDec.

% On the other hand, we find that EEM performs slightly worse than EnDec in terms of grammar and relevance. Because in the stage of inference, EnDec can generate the most possible responses without the limit of emotion. 

% To be mentioned, Emo-HRED shows the lowest emotion eliciting score. It proves that affect-sensitive models may be ineffective in positive-emotion elicitation when trained on general datasets.

% \noindent\textbf{Automatic Evaluation:} EEM also has the best performance on positive-emotion elicitation in automatic evaluation, with the highest estimated scores of $\hat{s}_2$ and $\Delta \hat{s}'$. 
% w/o CD and w/o CA are also more effective in emotion elicitation than EnDec and Emo-HRED.

\subsection{Case Study}
\label{sec:case}
We also provide an example of generated responses in Table~\ref{table5}. As can be seen, compared with the responses generated by EnDec and Emo-HRED, the responses generated by EEM with positive settings (\ie, $\lambda$s $>0.5$) are more user-concerned and achieve more positive emotion elicitation. Though some generated responses are not full of positive emotions themselves, they express a degree of empathy to effectively improve the emotional status of users. Besides, we present responses generated with different $\lambda$ (from 0.0 to 1.0 with an interval of 0.2).
As can be seen, EEM can generate responses with different categories of emotion elicitation, and all the responses are appropriate to the inputs. This implies the controllability and flexibility of EEM in emotion elicitation.

\textit{The implementation details, the distribution of emotion scores over the dataset, the visualization of the dual attention, and more cases are provided in Appendix.}

\vspace*{1.5mm}
\section{Conclusion and Future Work}
Endowing chatbots with the ability of positive emotion elicitation is valuable to improve the users' experience. However, manually collecting training labels for this task is labour-intensive, making existing solutions less practical. In this paper, we propose an emotion eliciting machine (EEM) trained on weak labels generated by a pretrained emotion classifier. We compute an emotion eliciting factor based on the weak labels, and propose a dual encoder-decoder structure to model the generation of responses in both positive and negative side based on the factor. 
% EEM is composed of a condition variable of emotion elicitation and a complementary encoder-decoder model.
% EEM uses two attention modules and two decoder modules to model the positive and negative emotion elicitation separately. A coefficient is then used to integrate both positive and negative modules. In this way, both positive and negative emotion elicitation can be learned more accurately, and the coefficient can be used to fine-grained control the degree of emotion elicitation.
Both the automatic and human evaluations on a large scale corpus show EEM outperforms the existing models in emotion eliciting conversation generation with better controllability and flexibility. In future work, we plan to refine EEM for multi-turn conversation and try to apply it on other text generation tasks.
% Moreover, we will study how the noise introduced by the sentiment classifier influence the generation. Finally, we believe that our approach is also promising in other tasks of , and corresponding experiments will be carried out for the validation.

\section*{Acknowledgement}
This work was supported by National Natural Science Foundation of China No. 61872370 and No. 61832017, Beijing Outstanding Young Scientist Program NO. BJJWZYJH012019100020098, and Shandong Provincial Natural Science Foundation under Grant ZR2019ZD06.

%\bibliographystyle{named}
%\bibliography{ijcai21}

\begin{thebibliography}{}

\bibitem[\protect\citeauthoryear{Baheti \bgroup \em et al.\egroup
  }{2018}]{baheti-etal-2018-generating}
A.~Baheti, A.~Ritter, J.~Li, and B.~Dolan.
\newblock Generating more interesting responses in neural conversation models
  with distributional constraints.
\newblock In {\em EMNLP}, 2018.

\bibitem[\protect\citeauthoryear{Colombo \bgroup \em et al.\egroup
  }{2019}]{colombo-etal-2019-affect}
P.~Colombo, W.~Witon, A.~Modi, J.~Kennedy, and M.~Kapadia.
\newblock Affect-driven dialog generation.
\newblock In {\em NAACL-HLT}, 2019.

\bibitem[\protect\citeauthoryear{Felbo \bgroup \em et al.\egroup
  }{2017}]{felbo-etal-2017-using}
B.~Felbo, A.~Mislove, A.~S{\o}gaard, I.~Rahwan, and S.~Lehmann.
\newblock Using millions of emoji occurrences to learn any-domain
  representations for detecting sentiment, emotion and sarcasm.
\newblock In {\em EMNLP}, 2017.

\bibitem[\protect\citeauthoryear{Fleiss}{1971}]{fleiss1971measuring}
J.~L. Fleiss.
\newblock Measuring nominal scale agreement among many raters.
\newblock {\em Psychological bulletin}, 76(5), 1971.

\bibitem[\protect\citeauthoryear{Hasegawa \bgroup \em et al.\egroup
  }{2013}]{hasegawa-etal-2013-predicting}
T.~Hasegawa, N.~Kaji, N.~Yoshinaga, and M.~Toyoda.
\newblock Predicting and eliciting addressee{'}s emotion in online dialogue.
\newblock In {\em ACL}, 2013.

\bibitem[\protect\citeauthoryear{Henderson \bgroup \em et al.\egroup
  }{2019}]{henderson-etal-2019-repository}
M.~Henderson, P.~Budzianowski, I.~Casanueva, S.~Coope, D.~Gerz, G.~Kumar,
  N.~Mrk{\v{s}}i{\'c}, G.~Spithourakis, P.-H. Su, I.~Vuli{\'c}, and T.-H. Wen.
\newblock A repository of conversational datasets.
\newblock In {\em Proceedings of the First Workshop on NLP for Conversational
  AI}, August 2019.

\bibitem[\protect\citeauthoryear{Huang \bgroup \em et al.\egroup
  }{2018}]{huang2018automatic}
C.~Huang, O.~Zaiane, A.~Trabelsi, and N.~Dziri.
\newblock Automatic dialogue generation with expressed emotions.
\newblock In {\em NAACL-HLT (Short Papers)}, 2018.

\bibitem[\protect\citeauthoryear{K. and
  B.}{2015}]{DBLP:journals/corr/KingmaB14}
Diederik~P. K. and Jimmy B.
\newblock Adam: {A} method for stochastic optimization.
\newblock In {\em ICLR}, 2015.

\bibitem[\protect\citeauthoryear{Li \bgroup \em et al.\egroup
  }{2015}]{li2015diversity}
J.~Li, M.~Galley, C.~Brockett, J.~Gao, and B.~Dolan.
\newblock A diversity-promoting objective function for neural conversation
  models.
\newblock {\em arXiv preprint arXiv:1510.03055}, 2015.

\bibitem[\protect\citeauthoryear{Lowe \bgroup \em et al.\egroup
  }{2015}]{lowe-etal-2015-ubuntu}
R.~Lowe, N.~Pow, I.~Serban, and J.~Pineau.
\newblock The {U}buntu dialogue corpus: A large dataset for research in
  unstructured multi-turn dialogue systems.
\newblock In {\em SIGdial}, 2015.

\bibitem[\protect\citeauthoryear{Lubis \bgroup \em et al.\egroup
  }{2018}]{lubis2018eliciting}
N.~Lubis, S.~Sakti, K.~Yoshino, and S.~Nakamura.
\newblock Eliciting positive emotion through affect-sensitive dialogue response
  generation: A neural network approach.
\newblock In {\em AAAI}, 2018.

\bibitem[\protect\citeauthoryear{Lubis \bgroup \em et al.\egroup
  }{2019}]{lubis2019positive}
N.~Lubis, S.~Sakti, K.~Yoshino, and S.~Nakamura.
\newblock Positive emotion elicitation in chat-based dialogue systems.
\newblock {\em TASLP}, 27(4), 2019.

\bibitem[\protect\citeauthoryear{Niu and Bansal}{2018}]{niu-bansal-2018-polite}
T.~Niu and M.~Bansal.
\newblock Polite dialogue generation without parallel data.
\newblock {\em TACL}, 6, 2018.

\bibitem[\protect\citeauthoryear{P. \bgroup \em et al.\egroup
  }{2019}]{DBLP:conf/nips/PaszkeGMLBCKLGA19}
Adam P., Sam G., Francisco M., Adam L., James B., Gregory C., Trevor K., Zeming
  L., Natalia G., Luca A., Alban D., Andreas K., Edward Y., Zachary D., Martin
  R., Alykhan T., Sasank C., Benoit S., Lu~F., Junjie B., and Soumith C.
\newblock Pytorch: An imperative style, high-performance deep learning library.
\newblock In {\em NeurIPS}, 2019.

\bibitem[\protect\citeauthoryear{Picard}{2000}]{picard2000affective}
R.~W. Picard.
\newblock {\em Affective computing}.
\newblock MIT press, 2000.

\bibitem[\protect\citeauthoryear{Pietquin and
  Hastie}{2013}]{Pietquin2013survey}
O.~Pietquin and H.~Hastie.
\newblock A survey on metrics for the evaluation of user simulations.
\newblock {\em The Knowledge Engineering Review}, 28(1), 2013.

\bibitem[\protect\citeauthoryear{Salovey and
  Mayer}{1990}]{salovey1990emotional}
P.~Salovey and J.~D. Mayer.
\newblock Emotional intelligence.
\newblock {\em Imagination, cognition and personality}, 9(3), 1990.

\bibitem[\protect\citeauthoryear{Serban \bgroup \em et al.\egroup
  }{2016}]{serban2016building}
I.~V. Serban, A.~Sordoni, Y.~Bengio, A.~Courville, and J.~Pineau.
\newblock Building end-to-end dialogue systems using generative hierarchical
  neural network models.
\newblock In {\em AAAI}, 2016.

\bibitem[\protect\citeauthoryear{Shantala \bgroup \em et al.\egroup
  }{2018}]{shantala2018neural}
R.~Shantala, G.~Kyselov, and A.~Kyselova.
\newblock Neural dialogue system with emotion embeddings.
\newblock In {\em SAIC}, 2018.

\bibitem[\protect\citeauthoryear{Shen and Feng}{2020}]{shen-feng-2020-cdl}
L.~Shen and Y.~Feng.
\newblock {CDL}: Curriculum dual learning for emotion-controllable response
  generation.
\newblock In {\em ACL}, 2020.

\bibitem[\protect\citeauthoryear{Song \bgroup \em et al.\egroup
  }{2019}]{song2019generating}
Z.~Song, X.~Zheng, Lu~Liu, M.~Xu, and X.-J. Huang.
\newblock Generating responses with a specific emotion in dialog.
\newblock In {\em ACL}, 2019.

\bibitem[\protect\citeauthoryear{Sordoni \bgroup \em et al.\egroup
  }{2015a}]{sordoni2015hierarchical}
A.~Sordoni, Y.~Bengio, H.~Vahabi, C.~Lioma, Jakob Grue~S., and J.-Y. Nie.
\newblock A hierarchical recurrent encoder-decoder for generative context-aware
  query suggestion.
\newblock In {\em CIKM}, 2015.

\bibitem[\protect\citeauthoryear{Sordoni \bgroup \em et al.\egroup
  }{2015b}]{sordoni2015neural}
A.~Sordoni, M.~Galley, M.~Auli, C.~Brockett, Y.~Ji, M.~Mitchell, J.-Y. Nie,
  J.~Gao, and B.~Dolan.
\newblock A neural network approach to context-sensitive generation of
  conversational responses.
\newblock {\em NAACL-HLT}, 2015.

\bibitem[\protect\citeauthoryear{Sutskever \bgroup \em et al.\egroup
  }{2014}]{sutskever2014sequence}
I.~Sutskever, O.~Vinyals, and Q.~V. Le.
\newblock Sequence to sequence learning with neural networks.
\newblock In {\em NeurIPS}, 2014.

\bibitem[\protect\citeauthoryear{Vaswani \bgroup \em et al.\egroup
  }{2017}]{vaswani2017attention}
A.~Vaswani, N.~Shazeer, N.~Parmar, J.~Uszkoreit, L.~Jones, A.~N Gomez, \L.
  Kaiser, and I.~Polosukhin.
\newblock Attention is all you need.
\newblock In {\em NeurIPS}, 2017.

\bibitem[\protect\citeauthoryear{Wu \bgroup \em et al.\egroup
  }{2017}]{wu-etal-2017-sequential}
Y.~Wu, W.~Wu, C.~Xing, M.~Zhou, and Z.~Li.
\newblock Sequential matching network: A new architecture for multi-turn
  response selection in retrieval-based chatbots.
\newblock In {\em ACL}, 2017.

\bibitem[\protect\citeauthoryear{Yuan \bgroup \em et al.\egroup
  }{2019}]{yuan-etal-2019-multi}
C.~Yuan, W.~Zhou, M.~Li, S.~Lv, F.~Zhu, J.~Han, and S.~Hu.
\newblock Multi-hop selector network for multi-turn response selection in
  retrieval-based chatbots.
\newblock In {\em EMNLP-IJCNLP}, 2019.

\bibitem[\protect\citeauthoryear{Zhou and Wang}{2018}]{zhou-wang-2018-mojitalk}
X.~Zhou and W.~Y. Wang.
\newblock {M}oji{T}alk: Generating emotional responses at scale.
\newblock In {\em ACL}, 2018.

\bibitem[\protect\citeauthoryear{Zhou \bgroup \em et al.\egroup
  }{2018}]{zhou2018emotional}
H.~Zhou, M.~Huang, T.~Zhang, X.~Zhu, and B.~Liu.
\newblock Emotional chatting machine: Emotional conversation generation with
  internal and external memory.
\newblock In {\em AAAI}, 2018.

\end{thebibliography}

\clearpage
\section{Appendix}
In this section, we provide the implementation of our method, the evaluation of the emotion classifier, the distribution of emotion scores over the dataset, the visualization of the dual attention mechanism, and more cases generated by each model.

\subsection{Implementation Details}
Our model is implemented by PyTorch~\cite{DBLP:conf/nips/PaszkeGMLBCKLGA19}. The encoders and decoders are both 2-layer GRUs with 600 hidden neurons in each layer. The word embedding dimension is set as 300, and the vocabulary size is set as 30,000. 
The trainable parameters are optimized by the Adam optimizer~\cite{DBLP:journals/corr/KingmaB14} with batch size and learning rate as 512 and 1e-4, respectively. All models are trained in five epochs.
% and the training stage of each model took 3 to 5 days on a P100 (16G) GPU machine. 
In the inference stage, beam search is applied for decoding responses with the beam size as five. 

\subsection{Evaluation of the Emotion Classifier}
200 utterances are randomly sampled for manual evaluation, and the emotion distribution is consistent with the whole dataset (the KL-divergence is 0.023). Each utterance is labeled as ``positive'', ``neutral'' or ``negative'' according to its emotion polarity by 5 annotators. Since the emotion score predicted by the classifier is continuous, we discrete it into ``negative'' (score between 0 and 0.35), ``neutral'' (score between 0.35 and 0.65), and ``positive'' (score between 0.65 and 1). As a result, the emotion classifier achieves 0.71 accuracy on average. The Fleiss' kappa coefficient~\cite{fleiss1971measuring} of the 5 annotations is 0.592, indicating ``Moderate agreement''.

\begin{figure}[t]
	\centering
	\includegraphics[width=0.99\linewidth,trim={2.5cm 10.5cm 3cm 10cm},clip]{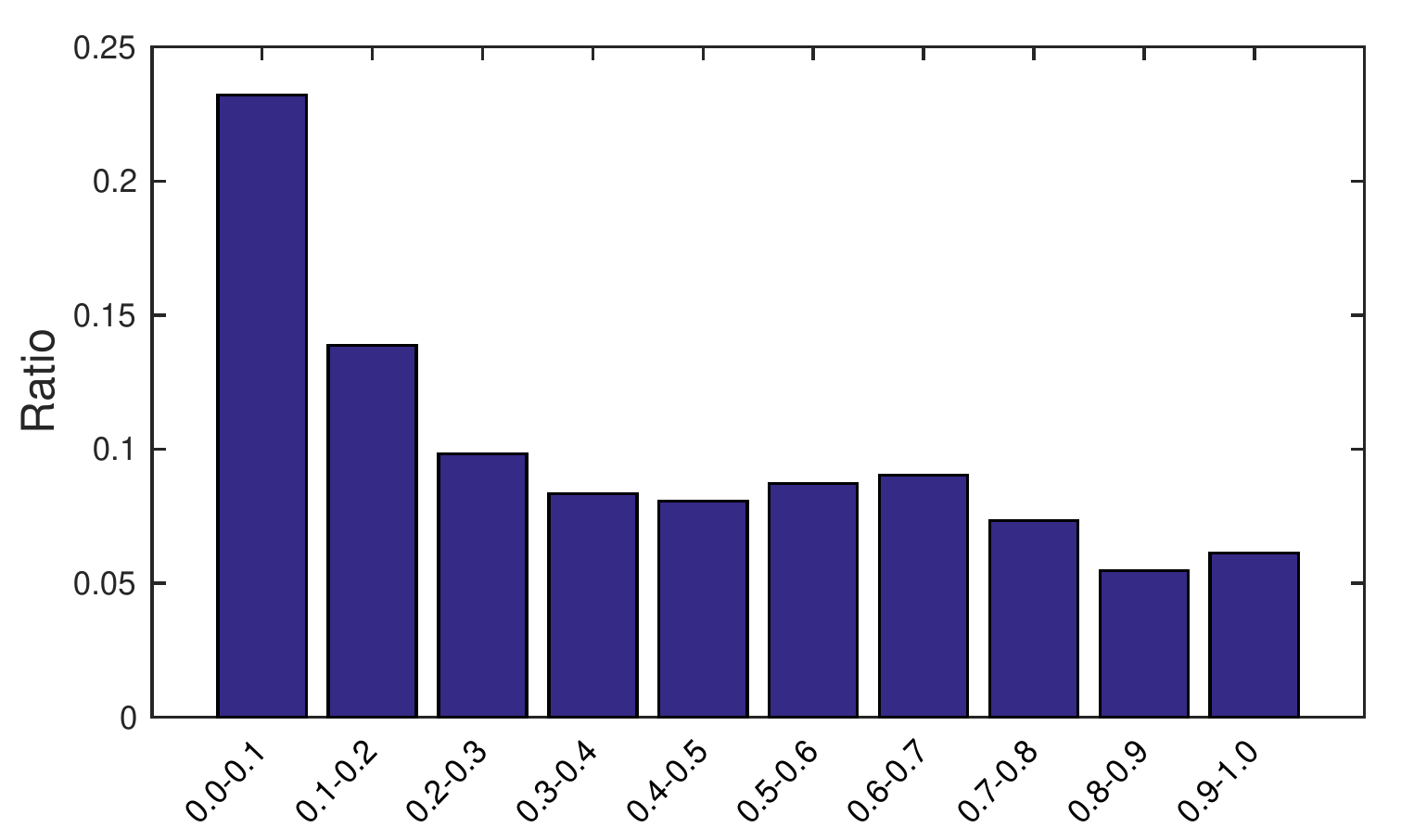}
	\caption{The distribution of the emotion scores over the Reddit dataset.}
	\label{ratio}
\end{figure}

\subsection{Distribution of Emotion Scores}
The distribution of the emotion scores is shown in Figure~\ref{ratio}. The ratio of the ``nagative'', ``neutral'', and ``positive'' data are around 47\%, 34\%, and 19\%, respectively. If we follow existing work~\cite{hasegawa-etal-2013-predicting,lubis2018eliciting,lubis2019positive} and generate binary labels for the data, we can only obtain 36.67\% data with positive emotion (\ie, the emotion score $>0.5$). In other words, more than 60\% data in the dataset cannot be used for training, which is a potential reason for their weak performance. 

\begin{figure}[t]
	\centering
	\includegraphics[width=\linewidth]{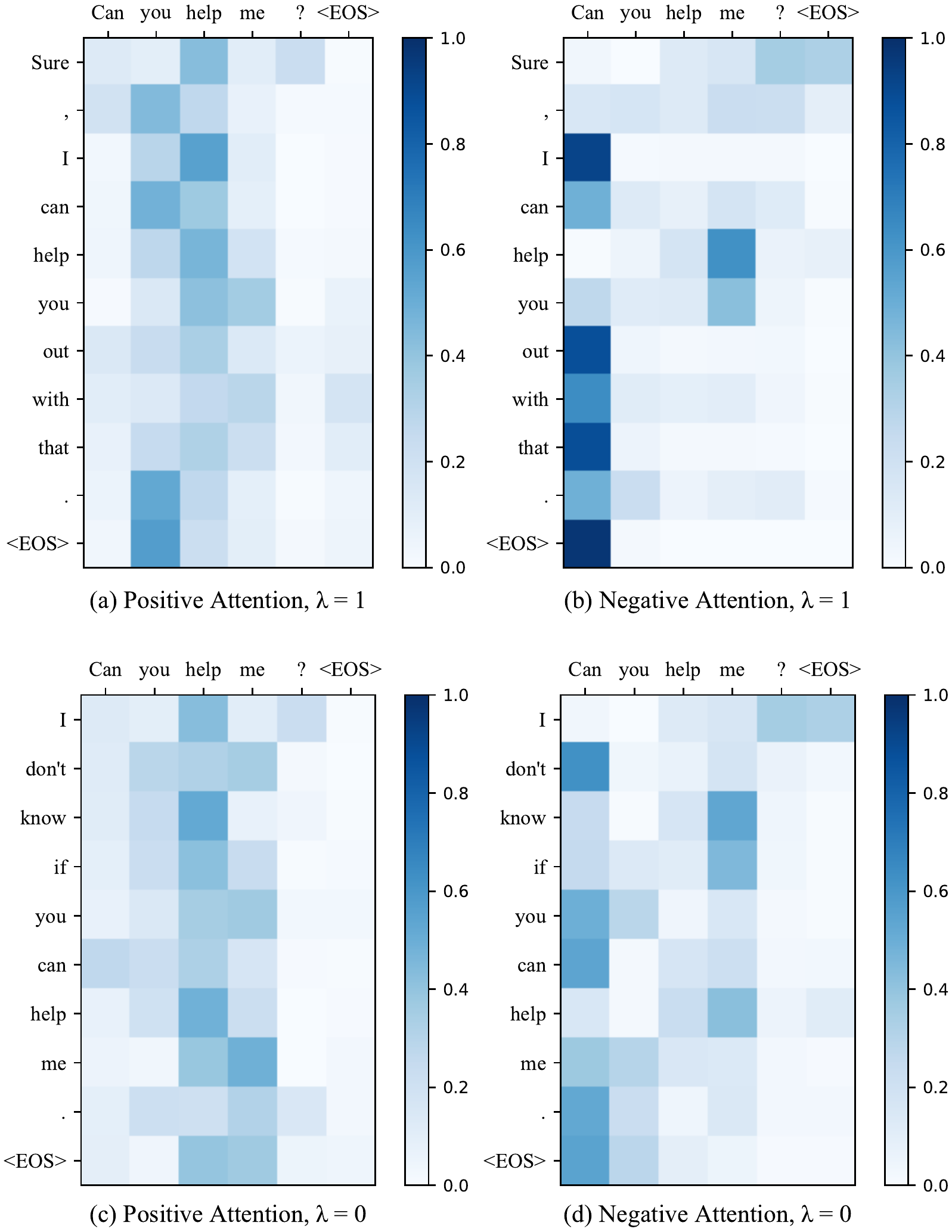} 
	\caption{Visualization of Dual Attention}
	\label{attention_visual}
\end{figure}

\subsection{Visualization of Attention}
To demonstrate the effectiveness and interpretability of the dual attention mechanism, we visualize the attention weights of the positive attention and the negative attention in the decoding process. The results are illustrated in Figure~\ref{attention_visual}. 

In this case, we input the message ``Can you help me?'' and set up different values of emotion eliciting factor $\lambda$ for EEM. As can be seen in subfigure (a) and (c), when $\lambda$ is set as 1.0, the positive attention is fully activated and it mainly focuses on ``help'', ``you'' and ``me'' in the input. The combination of these words implies the user's emotional demand (seeking help). Therefore, the activation of the positive attention contributes to generating an user-oriented and positive-emotion eliciting response as ``Sure, I can help you out with that''. 

On the other hand, when $\lambda$ is set as 0.0, the negative attention drives the decoder to focus on ``can'' and ``me'' (as shown in subfigure (b) and (d)), which implies rationality but more self-orientation and less relevance to the user's emotional status. As a result, the model will generate an emotionally suppressing response ``I don't know if you can help me''.

Based on these cases, we can see that our EEM can generate different responses according to the emotion eliciting factor $\lambda$ by activating different modules in the dual structure. Therefore, it is better at controlling the effect of positive emotion elicitation during the response generation process.

\subsection{Case Study}

\begin{table*}[t]
	\centering
	\small
	\begin{tabular}{p{3.5cm}|p{1.7cm}<{\centering}|p{1.7cm}<{\centering}|p{8.2cm}} %l(left)居左显示 r(right)居右显示 c居中显示
		\toprule
		\multicolumn{1}{p{3.0cm}|}{User Input}&\multicolumn{2}{c|}{Model}&\multicolumn{1}{c}{Response}\\
		\midrule
		\multirow{8}{3.0cm}{Can you help me?} & \multicolumn{2}{c|}{EnDec} & I can help you out.\\ 
		\cmidrule{2-4}
		& \multicolumn{2}{c|}{Emo-HRED} & I don't know what you're talking about.\\ 
		\cmidrule{2-4}
		& \multirow{6}{*}{\tabincell{c}{EEM}}&$\lambda=1.0$ & \textbf{Sure, I can help you out with that.}\\
		& & $\lambda=0.8$ & I love that game.\\ 
		& & $\lambda=0.6$ & I can help you out.\\ 
		& & $\lambda=0.4$ & I can help you out.\\ 
		& & $\lambda=0.2$ & I don't know what to do.\\ 
		& & $\lambda=0.0$ & I don't know if you can help me.\\
		\midrule
		\multirow{8}{3.0cm}{Let's go to play
		soccer together.} & \multicolumn{2}{c|}{EnDec} & I would love to play soccer.\\ 
		\cmidrule{2-4}
		& \multicolumn{2}{c|}{Emo-HRED} & What do you want?\\ 
		\cmidrule{2-4}
		& \multirow{6}{*}{\tabincell{c}{EEM}}&$\lambda=1.0$ & \textbf{I like to play with you!}\\
		& & $\lambda=0.8$ & Sure, I'll add you now.\\ 
		& & $\lambda=0.6$ & Let's do it.\\ 
		& & $\lambda=0.4$ & I'm not a fan of soccer.\\ 
		& & $\lambda=0.2$ & I don't know what to do.\\ 
		& & $\lambda=0.0$ & I don't know if you can help me.\\
		\bottomrule
	\end{tabular}
	%}
	\caption{Sample responses generated by EnDec, Emo-HRED, and EEM with different emotion eliciting factors.}
	\label{tab:case}
\end{table*}

In addtion to the case we shown in Section~\ref{sec:case}, we provide two more cases in Table~\ref{tab:case}. 
Similarly, EEM with $\lambda=1$ performs best on positive emotion elicitation. 

In particular, in the second case, EEM with $\lambda=1$ generates a response ``I like to play with you!''. Compared with responses generated by other models, tne personal pronoun ``you'' reduce the sense of distance between the chatbot and the user, thus can improve the user's emotion status directly. The responses generated by the baseline methods are fluent and naturally, but they have less effect on user's emotion (for example, ``I don't know what you're talking about'' generated by Emo-HRED in the first case). On the other hand, when $\lambda$ decreases from 1.0 to 0.0, it is evident that the emotion status of the generated response changed from positive to negative. This demonstrates that our proposed EEM can achieve a fine-grained controlling on positive emotion elicitation.

% the responses generated by EEM with $\lambda$ set to $1.0$ are most user-oriented and positive emotion eliciting. Though some generated responses are not full of positive emotions, they can effectively improve the emotional status of users.

% Additionally, we present responses generated with different $\lambda$s to illustrate the controllability and flexibility of EEM in emotion elicitation. As can be seen, EEM can generate responses with different categories of emotion elicitation, and all the responses are appropriate to the inputs.

\end{document}